\documentclass[10pt,twocolumn,letterpaper]{article}

\usepackage{cvpr}              
\usepackage[accsupp]{axessibility}  
\usepackage{marvosym}
\usepackage{cuted}  
\usepackage{tcolorbox}
\usepackage{url}

\definecolor{tabHeader}{HTML}{EFF6FF}   
\definecolor{tabStripe}{HTML}{FAFAFA}   
\definecolor{bestbg}{HTML}{DCFCE7}      
\definecolor{secondbg}{HTML}{FEF3C7}    
\definecolor{ourbg}{HTML}{E0F2FE}       
\definecolor{gainc}{HTML}{047857}       
\definecolor{lossc}{HTML}{B91C1C}       

\newcommand{\best}[1]{\cellcolor{bestbg}\textbf{#1}}
\newcommand{\secondbest}[1]{\cellcolor{secondbg}\underline{#1}}
\newcommand{\ourscell}[1]{\cellcolor{ourbg}\textbf{#1}}

\newcommand{\gainp}[1]{\textcolor{gainc}{\small\,(+#1\%)}}
\newcommand{\lossp}[1]{\textcolor{lossc}{\small\,(-#1\%)}}

\usepackage{booktabs}
\usepackage[table]{xcolor} 
\usepackage{makecell}
\definecolor{tabHeader}{HTML}{EFF6FF}   
\definecolor{tabHeader2}{HTML}{EDEDED}   
\definecolor{bioRedBG}{HTML}{FEE2E2} 
\definecolor{tabStripe}{HTML}{FAFAFA}   
\definecolor{bestbg}{HTML}{DCFCE7}      
\definecolor{secondbg}{HTML}{FEF3C7}    
\definecolor{ourbg}{HTML}{E0F2FE}       
\definecolor{gainc}{HTML}{047857}       
\definecolor{lossc}{HTML}{B91C1C}       

\definecolor{sectionRow}{RGB}{245,246,250}

\definecolor{cvprblue}{rgb}{0.21,0.49,0.74}
\usepackage[pagebackref,breaklinks,colorlinks,allcolors=cvprblue]{hyperref}
\newcommand{\Paragraph}[1]{\vspace{1mm} \noindent \textbf{#1} \hspace{0mm}}

\definecolor{tabHeader}{HTML}{EFF6FF}   
\definecolor{tabStripe}{HTML}{FAFAFA}   
\definecolor{bestbg}{HTML}{DCFCE7}      
\definecolor{secondbg}{HTML}{FEF3C7}    
\definecolor{ourbg}{HTML}{E0F2FE}       
\definecolor{gainc}{HTML}{047857}       
\definecolor{lossc}{HTML}{B91C1C}       

\renewcommand{\best}[1]{\cellcolor{bestbg}\textbf{#1}}
\renewcommand{\secondbest}[1]{\cellcolor{secondbg}\underline{#1}}
\renewcommand{\ourscell}[1]{\cellcolor{ourbg}\textbf{#1}}

\renewcommand{\gainp}[1]{\textcolor{gainc}{\small\,(+#1\%)}}
\renewcommand{\lossp}[1]{\textcolor{lossc}{\small\,(-#1\%)}}

\usepackage{booktabs}
\usepackage[table]{xcolor} 
\usepackage{makecell}
\definecolor{tabHeader}{HTML}{EFF6FF}   
\definecolor{tabHeader2}{HTML}{EDEDED}   
\definecolor{bioRedBG}{HTML}{FEE2E2} 
\definecolor{tabStripe}{HTML}{FAFAFA}   
\definecolor{bestbg}{HTML}{DCFCE7}      
\definecolor{secondbg}{HTML}{FEF3C7}    
\definecolor{ourbg}{HTML}{E0F2FE}       
\definecolor{gainc}{HTML}{047857}       
\definecolor{lossc}{HTML}{B91C1C}       
\usepackage{makecell}
\newcommand{\stackedmetric}[2]{\makecell[c]{#1\\#2}} 

\def\paperID{1649} 
\def\confName{CVPR}
\def\confYear{2026}

\title{From 3D Pose to Prose: Biomechanics-Grounded Vision--Language Coaching}

\author{
Yuyang Ji$^1$\quad
Yixuan Shen$^1$\quad
Shengjie Zhu$^2$\quad
Yu Kong$^2$\quad 
Feng Liu$^1$\textsuperscript{\Letter}\\
 $^1$ Department of Computer Science, Drexel University\\
  $^2$ Department of Computer Science and Engineering, Michigan State University\\
}

\usepackage{float}
\usepackage{dblfloatfix}

\begin{document}
\maketitle

\begingroup
\renewcommand{\thefootnote}{} 
\footnotetext{\Letter\ \texttt{fl397@drexel.edu}}
\endgroup

\vspace{-16mm}
\begin{strip}
\vspace{-20mm}
\centering
\includegraphics[width=\textwidth]{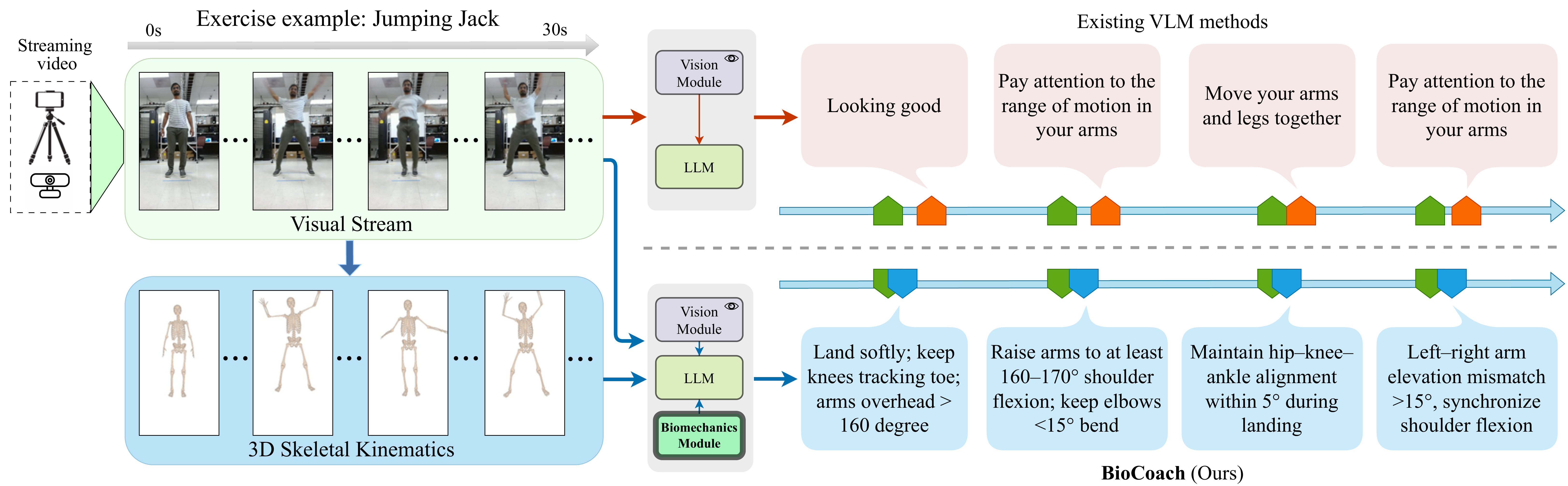}
\vspace{-6mm}
\captionof{figure}{Comparison with existing methods. Top: prior pixel-only VLM methods provide generic, loosely timed comments. Bottom: \textbf{BioCoach} fuses visual features with 3D skeletal kinematics and a biomechanics module to produce phase-aligned, anatomy-specific, quantitative cues (\emph{e.g.}, shoulder flexion 160$^\circ$–170$^\circ$), yielding more precise and biomechanics-grounded feedback along the same timeline.}
\vspace{-2mm}
\label{fig:teaser}
\end{strip}
\begin{abstract}
We present BioCoach, a biomechanics-grounded vision--language framework for fitness coaching from streaming video. BioCoach fuses visual appearance and 3D skeletal kinematics, through a novel three-stage pipeline: an exercise-specific degree-of-freedom selector that focuses analysis on salient joints; a structured biomechanical context that pairs individualized morphometrics with cycle and constraint analysis; and a vision--biomechanics conditioned feedback module that applies cross-attention to generate precise, actionable text. Using parameter-efficient training that freezes the vision and language backbones, BioCoach yields transparent, personalized reasoning rather than pattern matching. To enable learning and fair evaluation, we augment QEVD-fit-coach with biomechanics-oriented feedback to create QEVD-bio-fit-coach, and we introduce a biomechanics-aware LLM judge metric. BioCoach delivers clear gains on QEVD-bio-fit-coach across lexical and judgment metrics while maintaining temporal triggering; on the original QEVD-fit-coach, it improves text quality and correctness with near-parity timing, demonstrating that explicit kinematics and constraints are key to accurate, phase-aware coaching.
\href{https://vilab-group.com/project/biocoach}{Project} 
\end{abstract}    
\section{Introduction}
\label{sec:intro}

Live fitness coaching through streaming video is increasingly important, with a broad impact on at-home workouts~\cite{wilke2022train}, injury prevention~\cite{nilmart2024impact,jacobsson2023universal}, and data-driven rehabilitation~\cite{sumner2023artificial,abedi2024artificial,chen2026biogait}.
Traditionally, fitness coaching has relied on in-person trainers who provide immediate feedback to correct form deviations and reduce risks; however, expert coaching is expensive and often inaccessible. 
Advances in computer vision now enable automated form assessment from video~\cite{parmar2022domain,fieraru2021aifit}, yet existing approaches face three critical challenges that are also the field’s key requirements: (1) Timing and interaction, detecting coachable moments and delivering feedback at the right time rather than post hoc; (2) Biomechanical grounding, reasoning over 3D poses, joint angles, ranges of motion, and exercise phases instead of high-level appearance; and (3) Personalization and explainability, 
producing traceable, rule-level evidence.

Recent vision-language models have revitalized multimodal understanding~\cite{yin2024_mllm_survey,caffagni2024_mllm_survey,wang2024_mllm_review} and can generate fluent, instruction-like feedback~\cite{liu2023_llava,dai2023instructblip}; streaming VLMs such as Stream-VLM~\cite{panchal2024say} apply these capabilities to live fitness coaching. However, they remain largely prompt-driven and struggle to autonomously surface coachable moments; they lack explicit morphometric context for personalization; and they do not integrate symbolic biomechanical constraints, which can yield generic or mistimed advice (see Fig.~\ref{fig:teaser}).

To address these limitations, we propose \textbf{BioCoach}, a \emph{biomechanics-grounded vision-language} framework for streaming fitness coaching that bridges biomechanical analysis and multimodal understanding. Our core insight is to construct \emph{explicit, interpretable intermediate representations} that expose kinematic properties to the language model, enabling transparent reasoning about exercise form while preserving end-to-end trainability. Rather than treating visual appearance and 3D pose as disjoint streams or relying solely on pattern learning, BioCoach organizes them into a structured pipeline that grounds feedback in biomechanical principles and individual morphology (see Fig.~\ref{fig:teaser}).

Specifically, BioCoach extracts two complementary signals from streaming video: a visual stream that captures appearance and context, and a 3D kinematic stream that captures skeletal pose and body shape. The system then proceeds through three coordinated steps, each introducing a novel design. \emph{First}, an exercise-specific joint attention mechanism prioritizes the degrees of freedom relevant to the current exercise, focusing analysis on anatomically salient regions instead of treating all joints uniformly. \emph{Second}, a structured biomechanical context pairs morphometric information for personalization with cycle analysis against curated references to yield explicit, rule-based form cues, which replace implicit visual heuristics with verifiable evidence. \emph{Third}, a vision-biomechanics conditioning mechanism fuses visual evidence with this context to produce coaching that is temporally targeted, interpretable through intermediate representations, and grounded in biomechanical principles rather than relying solely on pattern matching.
Moreover, to support learning and evaluation, we create \emph{QEVD-bio-fit-coach} by re-annotating the QEVD-fit-coach dataset~\cite{panchal2024say} with fine-grained biomechanical annotations.

In summary, the contributions of this work include:

$\diamond$ We propose \textbf{BioCoach}, a new biomechanics-grounded vision-language framework for interpretable, personalized coaching from streaming video.


$\diamond$ We devise a three-stage pipeline unifying exercise-specific joint attention, biomechanical context for personalized rule-level cues, and vision–biomechanics conditioned feedback for targeted, auditable coaching.

$\diamond$ We create \textit{QEVD-bio-fit-coach} by re-annotating QEVD-fit-coach with fine-grained biomechanical labels.

$\diamond$ Extensive experiments show that \textbf{BioCoach} achieves superior text quality, timing accuracy, LLM-graded correctness, and biomechanics-grounded action quality scores.

\section{Related Work}
\label{sec:related_work}

\begin{figure*}[t]
  \centering
   \includegraphics[width=1\linewidth]{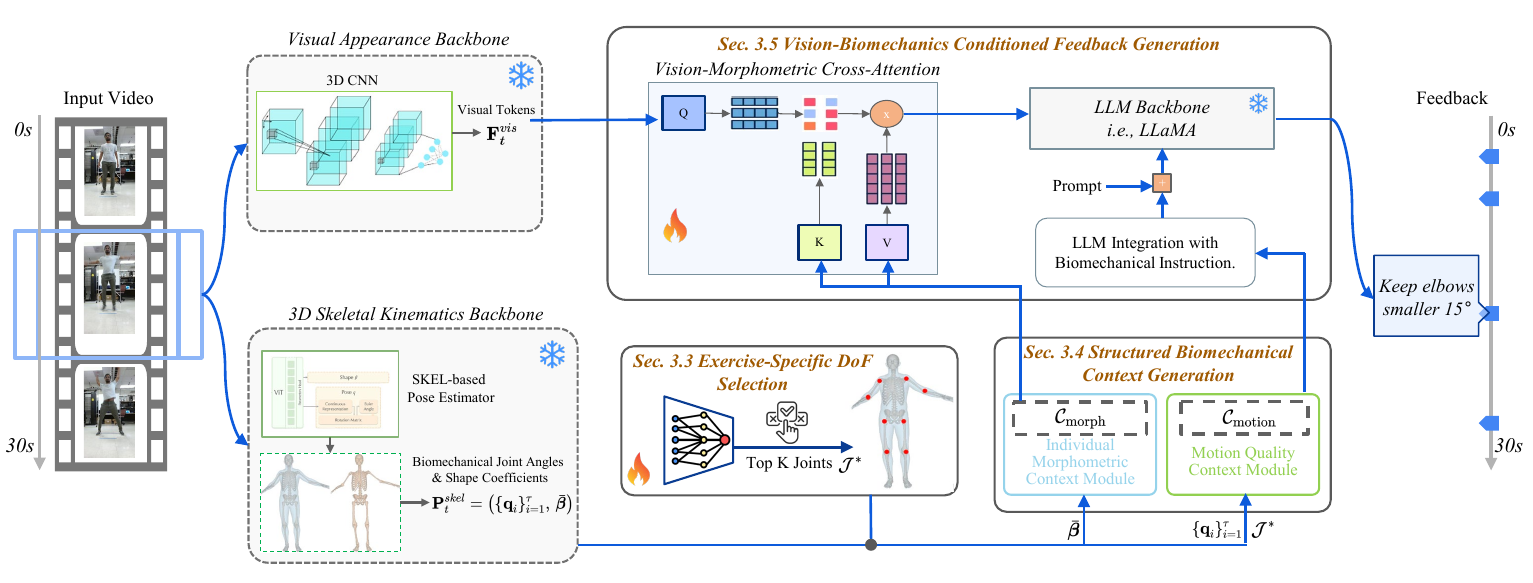}
   \vspace{-6mm}
\caption{\textbf{BioCoach overview.} Streaming video is encoded by two backbones: a 3D CNN for visual tokens and a pose extractor for 3D skeletal kinematics. The pipeline has three components: (1) Exercise-Specific DoF Selection uses a lightweight attention head to select the top $K$ biomechanically salient joints; (2) Structured Biomechanical Context builds two representations (individual morphometric context and motion quality context) capturing body measurements, cycles, ranges of motion, and constraint checks; (3) Vision--Biomechanics Conditioned Feedback fuses visual tokens with the morphometric context via cross-attention and prepends the motion-quality context as structured instruction to the LLM. This yields feedback grounded in explicit kinematic evidence rather than pattern matching alone.}
   \label{fig:flowchart}
   \vspace{0mm}
\end{figure*}

\Paragraph{Interactive Fitness Coaching and Vision-Language Models.}
Interactive fitness coaching from video has emerged as a key research direction with benchmarks like QEVD~\cite{panchal2024say} that pair long-range videos with timestamped coaching feedback. Recent vision-language models—including Flamingo~\cite{alayrac2022flamingo}, InstructBLIP~\cite{dai2023instructblip}, LLaVA variants~\cite{lin2024video,zhang2024llavanextvideo}, and LLaMA-VID~\cite{li2024llama}—demonstrate strong video understanding via gated cross-attention and token-level fusion. Stream-VLM~\cite{panchal2024say}, a strong QEVD baseline, introduced asynchronous interaction with action tokens to enable unprompted feedback. However, these models operate primarily on pixel-level features without explicit biomechanical constraints, leading to feedback that can be generic or poorly timed. BioCoach addresses this by extracting and integrating 3D skeletal kinematics as a structured modality, enabling biomechanics-grounded coaching.

\Paragraph{Pose, Shape Estimation, and Biomechanical Analysis.}
Advances in 3D human pose and mesh recovery span classic HMR~\cite{kanazawa2018end}, PARE~\cite{kocabas2021pare}, and CLIFF~\cite{li2022cliff}, and extend to recent SMPL-X–based methods that improve temporal stability and controllability, including PromptHMR~\cite{wang2025prompthmr} and joint-feature–guided HMR~\cite{yao2025jfghmr}, and CLIP-guided shape learning~\cite{liu2024distilling}, with contemporary surveys summarizing the state of the art~\cite{neupane2024survey,tian2023hmr_survey}. Complementary efforts push for biomechanically faithful modeling, such as SKEL~\cite{keller2023skin} and HSMR~\cite{xia2025reconstructing}, as well as gait models like Generative GaitNet~\cite{gaitnet} and Bidirectional GaitNet~\cite{gaitnetbidirectional} that condition on anatomical factors. Toolchains like OpenSim~\cite{delp2007opensim} and OpenCap~\cite{uhlrich2023opencap} estimate kinematics and even kinetics from video, enabling musculoskeletal analyses outside the lab. Prior fitness/AQA systems~\cite{pirsiavash2014assessing,parmar2019and,dwibedi2020counting,chen2022posetrainer} still tend to output scores or templates, analyze frames rather than full cycles, and ignore morphology~\cite{liu2023learning}.

\Paragraph{Structured Representations for Language Grounding.}
Structured interfaces for grounding language in perception span marker/slot prompts that tie text to localized evidence~\cite{yang2023set}, layout- or schema-driven planners that impose explicit spatial structure before generation~\cite{feng2024layoutgpt}, and visual in-context prompting that conditions models on exemplar prompts across tasks~\cite{li2024visualicp}. Recent “visual chain-of-thought” work equips multimodal LMs with sketchpads to externalize intermediate spatial reasoning~\cite{hu2024visualsketchpad}. In human motion, motion–language models discretize continuous kinematics into tokens to align motion and text within a single generative interface~\cite{jiang2023motiongpt}. Together, these directions show that explicit, compact structure can improve interpretability, controllability, and citation of evidence—key properties for corrective feedback on fine-grained physical behaviors.

\section{Method}\label{sec:method}

\subsection{Overview}\label{sec:overview}

The proposed \textbf{BioCoach} framework converts streaming fitness videos into biomechanically-grounded coaching feedback by creating \emph{explicit, interpretable intermediate representations} that bridge kinematic data and language generation. This design enables \emph{transparent reasoning} about exercise form, \emph{personalization} to individual body geometry, and \emph{feedback grounded in biomechanical principles} rather than learned patterns.
As illustrated in Fig.~\ref{fig:flowchart}, the framework extracts two complementary modalities from streaming video: visual appearance and 3D skeletal kinematics. These flow through a three-stage pipeline: (1) Exercise-Specific Degree-of-Freedom Selection Module (Sec.~\ref{sec:attention}) identifies anatomically salient joints; (2) Structured Biomechanical Context Generation Module (Sec.~\ref{sec:ebr}) analyzes motion quality while accounting for individual body geometry; (3) Vision-Biomechanics Conditioned Feedback Generation (Sec.~\ref{sec:coaching-module}) generates feedback grounded in explicit biomechanical analysis.

\subsection{Dual-Modality Feature Extraction}
\label{sec:dual-modality}

Our framework extracts two complementary modalities from streaming fitness 
videos. Both modalities originate from the video input $\mathbf{V} \in 
\mathbb{R}^{T \times H \times W \times 3}$ but capture different aspects 
of motion: visual appearance and 3D skeletal kinematics. To extract these 
modalities, we employ two specialized processing backbones, described below.

\Paragraph{Visual Appearance Backbone.}
Following~\cite{panchal2024say}, we employ a pre-trained 3D CNN that captures fine-grained motion patterns essential for fitness coaching (see \textbf{\emph{Supp}} for details). At each timestep $t$, the model processes a temporal window of video frames and extracts motion-aware features:
\begin{equation} 
\mathbf{F}_t^{vis} = \mathcal{F}(\mathbf{V}_{[t-\tau:t]}),
\end{equation} 
where $\mathcal{F}: \mathbb{R}^{\tau \times H \times W \times 3} \to \mathbb{R}^{N_v \times d}$ is the 3D CNN feature extractor, $\mathbf{V}_{[t-\tau:t]}$ denotes a sliding window of $\tau$ consecutive frames, and $\mathbf{F}_t^{vis} \in \mathbb{R}^{N_v \times d}$ contains $N_v$ visual feature tokens with an embedding dimension of $d$.
The architecture combines 2D and 3D convolutional layers: 2D layers capture spatial appearance cues within frames, while 3D layers encode temporal motion dynamics across the window. Crucially, all convolutions employ causal masking to ensure the model operates in a true streaming setting, generating predictions solely from past and current observations without access to future information.

\Paragraph{3D Skeletal Kinematic Backbone.}
%
Complementary to visual appearance features, we extract explicit 3D skeletal kinematics to capture biomechanically-grounded body motion. We employ HSMR~\cite{xia2025reconstructing} built upon SKEL~\cite{keller2023skin}, a frame-level method that produces pose estimates in biomechanically-grounded representations.
Following SKEL~\cite{keller2023skin}, we represent skeletal pose using 46-dimensional Euler-angle representations with joint-specific biomechanical constraints. To extract temporal kinematic information, we apply the pose extractor $\mathcal{P}$ to each frame in the temporal window $\mathbf{V}_{[t-\tau:t]}$:
\begin{equation}
\{\mathbf{q}_i\}_{i=1}^{\tau}, \{\boldsymbol{\beta}_i\}_{i=1}^{\tau} = 
\mathcal{P}\left(\mathbf{V}_{[t-\tau:t]}\right),
\end{equation}
where $\mathbf{q}_i \in \mathbb{R}^{46}$ represents biomechanical joint angles for frame $i$. To obtain a stable body shape representation across the temporal window, we aggregate per-frame shape coefficients via average pooling:
$\bar{\boldsymbol{\beta}} = \frac{1}{\tau}\sum_{i=1}^{\tau} \boldsymbol{\beta}_i$, 
expressed in SMPL-style parameterization. Thus, the kinematic modality output at time $t$ is:
$\mathbf{P}_t^{skel} = \left( \{\mathbf{q}_i\}_{i=1}^{\tau}, \, \bar{\boldsymbol{\beta}} \right)$,
where $\bar{\boldsymbol{\beta}} \in \mathbb{R}^{10}$ encodes individual body shape. Unlike appearance features that conflate body geometry with motion quality, skeletal kinematics provide normalized, biomechanically-aware representations that separate body shape from movement patterns.

\subsection{Exercise-Specific Degree-of-Freedom Selection}\label{sec:attention}
The biomechanical relevance of each joint depends on the exercise being performed. A squat requires lower-body joints (hips, knees, ankles), while a push-up demands upper-body joints (shoulders, elbows, wrists). This module identifies exercise-specific salient joints via visual context.

We leverage visual features $\mathbf{F}_t^{\text{vis}}$ to infer exercise-dependent joint relevance using a lightweight attention network $\mathcal{A}_\theta$:
\begin{equation}
\mathbf{s}^t = \mathcal{A}_\theta(\mathbf{F}_t^{\text{vis}}),
\end{equation}
where $\mathcal{A}_\theta$ is an MLP with learnable parameters $\theta$ that outputs importance scores for each joint. Each element $\mathbf{s}_j^t \in [0, 1]$ represents the biomechanical relevance of joint $j$. We select the top-$K$ joints by importance:
\begin{equation}
\mathcal{J}^* = \{j : \mathbf{s}_j^t \in \text{TopK}(\mathbf{s}^t, K)\}.
\end{equation}

Skeleton joints correspond to multiple degrees of freedom in the 46-dimensional Euler angle representation $\mathbf{q}$ (\eg, the shoulder has 3 degrees of freedom (DoFs): flexion/extension, abduction/adduction, and internal/external rotation). When a joint is selected in $\mathcal{J}^*$, all its associated DoFs are automatically included in downstream analysis, ensuring coherent joint-level reasoning. The focused joint set $\mathcal{J}^*$ remains fixed for the entire exercise session, mimicking how human coaches maintain consistent attention to relevant body regions throughout a movement.

\subsection{Structured Biomechanical Context Generation}
\label{sec:ebr}
This module generates subject-specific structured biomechanical contexts from 
kinematic data to enable personalized and biomechanically-grounded coaching. 
It consists of two submodules that, together, produce the individualized context required for the language module.

\subsubsection{Individual Morphometric Context Module}
\label{sec:morphometric}

Kinematic data must be contextualized within individual body geometry. This module extracts user-specific anthropometric measurements from the shape parameters $\bar{\boldsymbol{\beta}}$ of the kinematic modality $\mathbf{P}_t^{skel}$.
Raw shape parameters (\emph{i.e.}, SMPL~\cite{SMPL:2015} body shape coefficients $\bar{\boldsymbol{\beta}}$) are abstract and difficult for language models to interpret. To address this, we leverage Virtual Measurements~\cite{choutas2022accurate} to extract interpretable anthropometric measurements directly from the fitted SMPL mesh: \texttt{mass}, \texttt{height}, \texttt{chest}, \texttt{waist}, and \texttt{hip} circumference. Circumference measurements are derived via plane intersection, while length measurements use Euclidean distance between anatomical landmarks~\cite{choutas2022accurate}.
This grounds shape in physical quantities that are semantically meaningful for exercise assessment. The extracted measurements form the morphometric context $\mathcal{C}_{\text{morph}}$ which is formatted as human-readable descriptors:
\begin{tcolorbox}
[colback=gray!4,colframe=black,boxrule=0.4pt,arc=2mm,left=6pt,right=6pt,top=6pt,bottom=6pt]
\small
\texttt{User body: height $1.78\,\mathrm{m}$, mass $73.22\,\mathrm{kg}$, chest $1.00\,\mathrm{m}$, waist $0.83\,\mathrm{m}$, hips $0.98\,\mathrm{m}$.}
\end{tcolorbox}
\noindent By providing explicit, quantitative anchors grounded in individual body geometry, the morphometric context enables personalized and biomechanically-aware feedback. This representation is fused with visual features via cross-attention, allowing the language model to generate coaching feedback that accounts for individual differences.

 
\begin{figure}[t]
      \centering
   \includegraphics[width=1\linewidth]{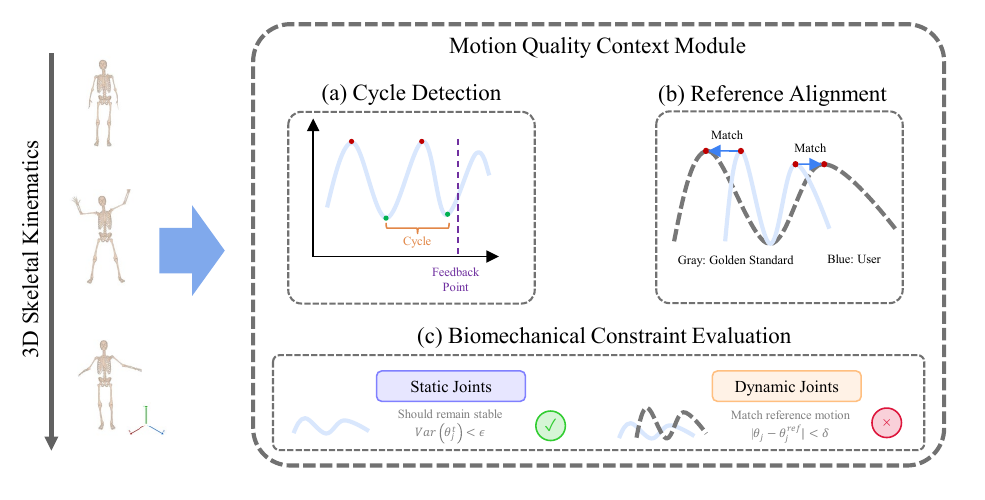}
   \caption{Motion-Quality Context module.  Given the selected joint set and the 3D skeletal kinematics, the module (a) detects repetition cycles and anchors the feedback moment; (b) time-normalizes each cycle and aligns it to a curated reference trajectory; and (c) evaluates biomechanical constraints: stability for static joints and deviation to reference for dynamic joints. Gray curves denote the reference, blue curves denote the user.}
   \label{fig:motion}
\end{figure}

\subsubsection{Motion Quality Context Module}
\label{sec:kce}

As illustrated in Fig.~\ref{fig:motion}, this module evaluates whether the user's motion satisfies biomechanical constraints for the given exercise, using the kinematic trajectory $\{\mathbf{q}_i\}_{i=1}^{\tau}$ and the selected joint set $\mathcal{J}^*$ from Sec.\ref{sec:attention}, and produces structured descriptions of motion quality and form deviations.

\Paragraph{Cycle detection.} 
Motion cycles are identified by analyzing key joints in $\mathcal{J}^*$. 
At each frame $i$ in the session, we have the full angle vector $\mathbf{q}_i \in \mathbb{R}^{46}$; 
we denote the primary DoF angle of joint $j \in \mathcal{J}^*$ at frame $i$ as $q_{j,i}^*$, 
where $q_{j,i}^* = q_{d(j),i}$ and $d(j)$ maps joint $j$ to its primary DoF index in $\mathbf{q}_i$ 
(\emph{e.g.}, $d(\text{knee}) = 7$ for knee flexion).
For repetitive exercises (\emph{e.g.}, squats, push-ups), we select a representative 
joint $j$ and extract its angle trajectory $\{q_{j,i}\}_{i=1}^{N}$ 
over the entire session. After gaussian smoothing, we identify cycle peaks and valleys 
using prominence-based peak detection, producing candidate cycle boundaries $(i_s, i_e)$. 
Spurious detections are filtered by excluding cycles that are shorter than $\tau_{\min}$ 
or longer than $\tau_{\max}$.
For exercise-specific variants: alternating movements (\emph{e.g.}, jumping jacks) use zero-crossing detection within the detected cycle. 
We compute the cycle-specific mean angle $\bar{q}_{j,\text{cycle}} = \frac{1}{i_e - i_s + 1}\sum_{i=i_s}^{i_e} q_{j,i}$ 
and detect zero-crossings of the normalized signal $q_{j,i} - \bar{q}_{j,\text{cycle}}$. 
Static holds (\emph{e.g.}, planks) identify low-variance regions where 
$\text{Var}(\{q_{j,i}\}_{i \in [i_s, i_e]}) < \epsilon$, indicating that the joint remains relatively stable within the detected cycle.

\Paragraph{Reference Alignment.} 
Each detected cycle $(i_s, i_e)$ is temporally normalized by resampling to match a curated reference trajectory $\mathbf{R}$ via linear interpolation. For joint $j$, we interpolate sequences:
\begin{equation}
\tilde{q}_{j,k} = q_{j, \lfloor \phi(k) \rfloor} + (\phi(k) - 
\lfloor \phi(k) \rfloor)(q_{j, \lceil \phi(k) \rceil} - 
q_{j, \lfloor \phi(k) \rfloor}),
\end{equation}
where $\phi(k)$ maps the reference index $k \in [1, N_{\text{ref}}]$ to the user's cycle timeline $i \in [i_s, i_e]$, and $N_{\text{ref}}$ is the reference cycle length. Function $\phi$ rescales cycles to align with the reference length, ensuring that both trajectories map to a common temporal grid. The subscript $j$ indicates the scalar angle trajectory for joint $j$, not the full 46-dimensional vector.
We compute similarity scores by comparing the interpolated user cycle $\{\tilde{q}_{j,k}\}$ with the reference trajectory, which includes (1) cosine similarity on normalized angle profiles, (2) joint-specific Pearson correlations, (3) temporal derivative alignment measuring velocity consistency, and (4) amplitude comparison of the range of motion. These metrics yield a cycle quality score $s_{\text{cycle}} \in [0,1]$.

\Paragraph{Biomechanical Constraint Evaluation.}
For joint $j \in \mathcal{J}^*$, we evaluate exercise-specific biomechanical constraints based on the detected cycle $(i_s, i_e)$. Each joint is classified as \emph{static} (should remain stable) or \emph{dynamic} (should exhibit specific motion patterns), as determined by domain experts or LLM-assisted annotation (see \textbf{\emph{Supp}} for details). For static joints, we measure stability via variance:
$\delta_j^{\text{static}} = \text{Var}(\{q_{j,i}\}_{i \in [i_s, i_e]})$
For dynamic joints, we measure deviation from the reference trajectory at critical frames:
$
\delta_j^{\text{dynamic}} = |q_j^{\text{user}}(i_{\text{key}}) - q_j^{\text{ref}}(i_{\text{key}})|,
$
where $i_{\text{key}}$ denotes critical frames (\emph{e.g.}, the deepest point of a squat).
Violations are detected by checking against exercise-specific acceptable bounds $[l_j, u_j]$:
\begin{equation}
\text{violation}_j = \begin{cases}
1, & \text{if } \delta_j < l_j \text{ or } \delta_j > u_j \\
0, & \text{otherwise}.
\end{cases}
\end{equation}
When detected ($\text{violation}_j = 1$), we generate descriptions quantifying deviations and corrections.

\Paragraph{Structured Prompt Generation.}
At each frame, the module produces two outputs. Pose state captures selected joint configurations:
\begin{equation}
\mathbf{p}_{\text{state}}^{(i)} = \{\texttt{joint}_{j}\text{ angle } 
\lfloor q_{j,i} \rfloor^{\circ} \mid j \in \mathcal{J}^*\},
\end{equation}
For joints where violations are detected ($\text{violation}_j = 1$), we generate descriptions quantifying deviations:
$\mathbf{v}_{\text{violations}} = \{\text{``}\texttt{joint}_j\text{ deviation: } 
\delta_j\text{ (bound: } [l_j, u_j]\text{)''} \mid \text{violation}_j = 1\}$. 
These are formatted as natural language:
 \begin{tcolorbox}[colback=gray!4,colframe=black,boxrule=0.4pt,arc=2mm,left=6pt,right=6pt,top=6pt,bottom=6pt]
 \small
\texttt{Current pose: right knee $85^{\circ}$, left knee $88^{\circ}$, hip $75^{\circ}$. \\
Form issues: Right knee flexion insufficient ($85^{\circ}$ detected, $90^{\circ}$ required)}
\end{tcolorbox}
\noindent Motion quality context concatenates pose state and violations:
\begin{equation}
\mathcal{C}_{\text{motion}} = [\mathbf{p}_{\text{state}}^{(i)}; \mathbf{v}_{\text{violations}}],
\end{equation}
which enables precise feedback grounded in biomechanical analysis.


\subsection{Vision-Biomechanics Conditioned Feedback Generation}
\label{sec:coaching-module}
This module synthesizes coaching feedback by conditioning a language model (\emph{i.e.}, LLaMA-2-7B~\cite{touvron2023llama}) on visual features $\mathbf{F}_t^{\text{vis}}$, morphometric context $\mathcal{C}_{\text{morph}}$, and motion quality context $\mathcal{C}_{\text{motion}}$.

\Paragraph{Structured Multimodal Context Encoding.}
Both morphometric and motion contexts are encoded into token embeddings through 
the language model's embedding layer:
\begin{equation}
\mathbf{m}_t = \text{Embed}(\mathcal{C}_{\text{morph}}), \quad
\mathbf{c}_t = \text{Embed}(\mathcal{C}_{\text{motion}}),
\end{equation}
where $\mathbf{m}_t $ encodes body measurements and pose, 
and $\mathbf{c}_t $ encodes motion quality and constraint 
violations.

\Paragraph{Vision-Morphometric Cross-Attention.}
%
%
Visual features $\mathbf{F}_t^{\text{vis}}$ are grounded in 
morphometric tokens $\mathbf{m}_t$ through cross-attention:
\begin{equation}
\mathbf{z}_t = \mathbf{F}_t^{\text{vis}} + \text{CrossAttn}(\mathbf{F}_t^{\text{vis}}, \mathbf{m}_t, \mathbf{m}_t),
\end{equation}
where
\begin{equation}
\resizebox{\linewidth}{!}{$
\operatorname{CrossAttn}\!\left(\mathbf{F}_t^{\text{vis}}, \mathbf{m}_t, \mathbf{m}_t\right)
= \operatorname{Softmax}\!\left(
\frac{\mathbf{F}_t^{\text{vis}} W_Q (\mathbf{m}_t W_K)^\top}{\sqrt{d}}
\right)\cdot (\mathbf{m}_t W_V)
$},
\label{eq:crossattn}
\end{equation}
with $W_Q, W_K, W_V$ being learnable projection matrices. This produces fused visual-morphometric features $\mathbf{z}_t$ that align visual observations with individual body geometry.

\Paragraph{LLM Integration with Biomechanical Instruction.}
Motion quality context $\mathcal{C}_{\text{motion}}$ encapsulates rich biomechanical 
analysis, including cycle-based motion quality, constraint violations, 
and quantified form deviations. This context is prepended as structured instruction to directly guide 
language generation with explicit biomechanical constraints:
\begin{equation}
\text{Prompt} = [\text{Embed}(\mathcal{C}_{\text{motion}}), \text{language\_tokens}].
\end{equation}
The LLM then generates feedback conditioned on this explicit biomechanical analysis:
\begin{equation}
\text{Feedback}_t = \text{LLM}(\text{Prompt}. \{\mathbf{z}_t\}).
\end{equation}
By injecting structured biomechanical constraints directly into the prompt rather 
than relying solely on learned patterns, this design ensures that feedback is grounded 
in explicit, exercise-specific biomechanical principles. (See \textbf{\emph{Supp}} for mathematical details.)

\subsection{Training Objective}
\label{sec:Training-Objective}
We employ a parameter-efficient fine-tuning strategy that freezes both the 3D CNN visual backbone and the LLaMA-2-7B language model, updating only the cross-attention fusion layers and the DoF selection network $\mathcal{A}_\theta$. This design preserves pre-trained linguistic priors while adapting multimodal fusion to align visual appearance, skeletal kinematics, and coaching feedback. 

\Paragraph{Exercise-Specific DoF Selection Training.}
The attention network $\mathcal{A}_\theta$ is trained with annotated exercise-specific joint relevance labels from domain experts and LLM-assisted annotation. We use binary cross-entropy loss to align predicted importance scores $\mathbf{s}^t$ with ground-truth salient joint sets $\mathcal{J}^{\text{gt}}$:
\begin{equation}
    \mathcal{L}_{\text{DoF}} = -\sum_{j=1}^{J} \left[ y_j \log(\mathbf{s}_j^t) + (1-y_j) \log(1-\mathbf{s}_j^t) \right],
\end{equation}
where $y_j = 1$ for biomechanically-relevant joints and $y_j = 0$ otherwise.

\Paragraph{Cross-Attention Fusion.}
Training mirrors the streaming inference regime using autoregressive cross-entropy loss with selective down-weighting to prevent excessive deferral of feedback:
\begin{equation}
    \mathcal{L}_{\text{CE}} = -\sum_{t=1}^{N-1} w_{x_{t+1}} \log P(x_{t+1} \mid x_{\le t}),
\end{equation}
where $w_{x_{t+1}} = \alpha < 1$ for continuation tokens (\texttt{<next>} action tokens) and $w_{x_{t+1}} = 1$ for feedback content. This down-weighting encourages timely feedback generation rather than indefinite observation windows.

\subsection{QEVD-bio-fit-coach Dataset}
\label{sec:qevid}
We augment QEVD-fit-coach~\cite{panchal2024say} (149 training videos, 74 test videos, 23 exercises, $\sim$2,484 timestamped feedback) to create \textbf{QEVD-bio-fit-coach}. The enhanced set adds biomechanically grounded feedback annotations. Using these signals, we systematically rewrite colloquial feedback into anatomically precise language: generic cues like ``lower your body more'' become quantified targets (``increase elbow flexion to $90^{\circ}$ at the bottom''), and vague instructions become explicit alignments (``maintain a neutral spine, head to heels''). We also add brief biomechanical rationales (\eg, ``increase hip/knee flexion to distribute load''). The temporal boundaries of all feedback instances are preserved exactly as in the original; only the \emph{content} is rendered more technical. This controlled augmentation isolates the effect of explicit biomechanical terminology and enables evaluation of whether kinematic grounding improves coaching quality.

\subsection{Implementation Details}
\label{sec:impl}
Following Stream-VLM~\cite{panchal2024say}, we adopt LLaMA-2-7B~\cite{touvron2023llama} as the language backbone and a pre-trained 3D CNN for visual features. Skeletal kinematics are represented by 46-dimensional Euler angles with a temporal window of 3 seconds ($\tau=12$) of motion history to detect exercise cycles and evaluate biomechanical constraints.
The DoF-selection network $\mathcal{A}_\theta$ is a 3-layer MLP (ReLU activations, sigmoid output) that produces per-joint importance scores $\mathbf{s}^t\in[0,1]^J$; we select the top $K=12$ salient joints per exercise.
We train two variants: one on \emph{QEVD-fit-coach} (original feedback) and one on \emph{QEVD-bio-fit-coach} (biomechanics-grounded feedback). Both fine-tune only the cross-attention layers and $\mathcal{A}_\theta$ using AdamW (learning rate $2\times10^{-5}$), a batch size of 8, and action-token down-weighting $\alpha=0.1$. DoF selection is supervised using a balanced binary cross-entropy loss.
For biomechanical analysis, cycle durations are constrained to $[\tau_{\min},\tau_{\max}]=[0.8,5]\,\mathrm{s}$. Constraint thresholds are exercise-specific: static-joint variance $<5^{\circ}$; dynamic-joint deviation tolerance $\pm5^{\circ}$ to $\pm10^{\circ}$ at key frames. Videos are resized to $H=224,W=160$ with standard augmentations. (See \textbf{\emph{Supp}} for additional implementation details.)


\begin{table*}[t]
\centering
\caption{Evaluation on the \emph{QEVD-bio-fit-coach}, a newly created benchmark with fine-grained biomechanical ground-truth feedback annotations. Second line in each cell shows \% improvement vs.\ Stream-VLM~\cite{panchal2024say}. \emph{LLM-Bio-Acc.} is our LLM-as-judge metric tailored to biomechanics, assessing the biomechanical correctness and specificity of generated feedback.}
\setlength{\tabcolsep}{10pt}
\small
{\renewcommand{\arraystretch}{1.3}
\resizebox{1\linewidth}{!}{
\begin{tabular}{lcccccc}
\rowcolor{tabHeader2}
\toprule
\textbf{Method} & \textbf{METEOR} $\uparrow$ & \textbf{ROUGE-L} $\uparrow$ & \textbf{BERTScore} $\uparrow$  & \textbf{LLM-Acc.} $\uparrow$   & \cellcolor{bioRedBG}\textbf{LLM-Bio-Acc.} $\uparrow$ & \textbf{T-F-Score} $\uparrow$ \\
\midrule
\stackedmetric{Stream-VLM~\cite{panchal2024say}}{(NeurIPS '24) } & 0.086 & 0.108 & 0.852 & 1.86 & 1.72 & 0.530 \\
\textbf{BioCoach} &
\stackedmetric{\ourscell{0.312}}{\gainp{262.8}} &
\stackedmetric{\ourscell{0.302}}{\gainp{179.6}} &
\stackedmetric{\ourscell{0.877}}{\gainp{2.9}}   &
\stackedmetric{\ourscell{3.12}}{\gainp{67.7}}   &
\stackedmetric{\ourscell{3.26}}{\gainp{89.5}}   &
\stackedmetric{\ourscell{0.544}}{\gainp{2.6}}   \\
\bottomrule
\end{tabular}
\label{tab:results_enhanced}
}
}
\end{table*}

\begin{table*}[t]
\centering
\caption{Performance on the QEVD-fit-coach with original feedback annotations. Parenthesized values are \% change vs. the best baseline (Stream-VLM~\cite{panchal2024say}). $\dagger$ zero-shot without fine-tuning.}
\setlength{\tabcolsep}{6pt}
\rowcolors{3}{tabStripe}{white}
\resizebox{0.92\linewidth}{!}{
\begin{tabular}{lccccc}
\rowcolor{tabHeader2}
\toprule
\textbf{Method} & \textbf{METEOR} $\uparrow$ & \textbf{ROUGE-L} $\uparrow$ & \textbf{BERTScore} $\uparrow$ & \textbf{LLM-Acc.} $\uparrow$ & \textbf{T-F-Score} $\uparrow$ \\
\midrule
\multicolumn{6}{l}{\textit{Zero-shot Models}} \\
InstructBLIP$^\dagger$~\cite{dai2023instructblip}   & 0.047 & 0.040 & 0.839 & 1.56 & - \\
Video-LLaVA$^\dagger$~\cite{lin2024video}    & 0.057 & 0.025 & 0.847 & 2.16 & - \\
Video-ChatGPT$^\dagger$~\cite{maaz2024video}  & 0.098 & 0.078 & 0.850 & 1.91 & - \\
Video-LLaMA$^\dagger$~\cite{zhang-etal-2023-video}    & 0.101 & 0.077 & 0.859 & 1.29 & - \\
LLaMA-VID$^\dagger$~\cite{li2024llama}      & 0.100 & 0.079 & 0.859 & 2.20 & - \\
LLaVA-NeXT$^\dagger$~\cite{zhang2024llavanextvideo}     & 0.104 & 0.078 & 0.858 & 2.27 & - \\
\midrule
\multicolumn{6}{l}{\textit{Fine-tuned Models}} \\
Socratic-LLaMA-2-7B~\cite{touvron2023llama}      & 0.094 & 0.071 & 0.860 & 2.17 & 0.50 \\
Video-ChatGPT~\cite{maaz2024video}            & 0.108 & 0.093 & 0.863 & 2.33 & 0.50 \\
LLaMA-VID~\cite{li2024llama}                & 0.106 & 0.090 & 0.860 & 2.30 & 0.50 \\
Stream-VLM~\cite{panchal2024say} (NeurIPS '24) & \secondbest{0.127} & \secondbest{0.112} & \secondbest{0.863} & \secondbest{2.45} & \best{0.56} \\
\textbf{BioCoach} & \ourscell{0.129}\gainp{1.6} & \ourscell{0.122}\gainp{8.9} & \ourscell{0.864}\gainp{0.1} & \ourscell{2.56}\gainp{4.5} & \ourscell{0.544}\lossp{2.9} \\
\bottomrule
\end{tabular}
}
\label{tab:results_original}
\end{table*}

\begin{table*}[t]
\centering
\small
\setlength{\tabcolsep}{6pt}
\rowcolors{3}{tabStripe}{white}
\caption{Ablation study on \emph{QEVD-Bio-Fit-Coach}. Each row removes or modifies one component while keeping others fixed.}
\resizebox{1\linewidth}{!}{
{\renewcommand{\arraystretch}{1.25}
\begin{tabular}{lcccccc}
\rowcolor{tabHeader2}
\toprule
\textbf{Model Variant} & \textbf{METEOR} $\uparrow$ & \textbf{ROUGE-L} $\uparrow$ &
\textbf{BERTScore} $\uparrow$ & \textbf{LLM-Acc.} $\uparrow$ &
\cellcolor{bioRedBG}\textbf{LLM-Bio-Acc.} $\uparrow$ & \textbf{T-F-Score} $\uparrow$  \\
\midrule
\textbf{Full Model} ($\tau = 3\,\mathrm{s}$ ) &
\ourscell{0.312} & \ourscell{0.302} & \ourscell{0.877} &
\ourscell{3.12} & \ourscell{3.26} & \ourscell{0.544} \\
\midrule
w/o Exercise-Specific DoF Selection &
0.305 & 0.296 & 0.873 & 3.06 & 3.14 & 0.543 \\
w/o Motion Quality Context Module &
0.133 & 0.127 & 0.864 & 2.63 & 2.04 & 0.544 \\
w/o Individual Morph Context Module &
0.284 & 0.273 & 0.871 & 2.91 & 3.07 & 0.535 \\
Window size $\tau = 2\,\mathrm{s}$ &
0.311 & 0.303 & 0.876 & 3.10 & 3.22 & 0.416 \\
\bottomrule
\end{tabular}
}}
\label{tab:ablation}
\end{table*}

\section{Experiments}\label{sec:experiments}

\subsection{Experimental Setup}

\Paragraph{Dataset.}
We evaluate on QEVD-fit-coach~\cite{panchal2024say}, a streaming fitness-coaching benchmark with 74 test videos across 23 exercises and $\sim$2{,}484 timestamped feedback instances. To better assess biomechanical understanding, we also report results on the \emph{QEVD-bio-fit-coach}.


\Paragraph{Baselines:}
We compare BioCoach against state-of-the-art vision–language models in two settings:
\begin{itemize}[leftmargin=*,itemsep=2pt]
\item Zero-shot models: InstructBLIP~\cite{dai2023instructblip}, Video-LLaVA~\cite{lin2024video}, Video-ChatGPT~\cite{maaz2024video}, Video-LLaMA~\cite{zhang-etal-2023-video}, LLaMA-VID~\cite{li2024llama}, and LLaVA-NeXT~\cite{zhang2024llavanextvideo}. These models are evaluated without any fitness-domain tuning and are periodically prompted to generate feedback from the video stream.

\item Fine-tuned models: Methods trained on the QEVD-fit-coach benchmark~\cite{panchal2024say}:
\begin{itemize}[leftmargin=*,itemsep=1pt]
    \item Socratic-LLaMA-2-7B~\cite{panchal2024say}: A text-only baseline that feeds activity descriptors from a fine-tuned 3D CNN into LLaMA-2-7B to produce feedback.
    \item Video-ChatGPT~\cite{maaz2024video} and LLaMA-VID~\cite{li2024llama}: Turn-based VLMs fine-tuned on QEVD-FIT-COACH that use CLIP/ViT visual encoders and require prompting at fixed intervals.
    \item Stream-VLM~\cite{panchal2024say}: The original streaming baseline on QEVD-FIT-COACH, combining a 3D CNN vision encoder with a LLaMA-2 language model and special \texttt{<next>}/\texttt{<feedback>} tokens for interactive, timely responses. This is the strongest, fine-tuned baseline tailored for fitness coaching.
\end{itemize}
\end{itemize}

\Paragraph{Metrics.}
Following Stream-VLM~\cite{panchal2024say}, we report:
\begin{itemize}[leftmargin=*,itemsep=2pt]
\item METEOR~\cite{banerjee2005meteor}: lexical overlap with stemming, synonym, and paraphrase matching. Rewards correct technical terms and entities (\emph{e.g.}, if the reference says ``arm not moving,'' predictions should mention the arm).
\item ROUGE-L~\cite{lin2004rouge}: longest–common–subsequence overlap, capturing phrase ordering and multiword expression similarity.
\item BERTScore~\cite{zhangbertscore}: semantic similarity via contextual embeddings and cosine similarity, robust to paraphrasing beyond exact word overlap.
\item LLM-Accuracy~\cite{panchal2024say}: LLaMA-3-70B-Instruct~\cite{dubey2024llama} acts as an automatic judge to holistically rate each feedback on a 1–5 scale (higher is better), considering context, correctness, and appropriateness; this correlates better with human preferences than purely lexical metrics.
\item Temporal F-Score (T-F-Score)~\cite{panchal2024say}: time-alignment precision/recall $F_1$ indicating whether feedback is issued at the correct moments.
\end{itemize}
In addition, to evaluate biomechanical fidelity on QEVD-bio-fit-coach, we introduce LLM-Bio-Accuracy, an LLM-as-judge metric specialized for biomechanics. Using the same LLaMA-3-70B-Instruct with a domain-specific prompt, it scores (1--5) factual biomechanical correctness and relevance against the reference context (implicitly reflecting per-frame kinematics, phase labels, and constraint annotations), emphasizing precise anatomy (joint names/sides), angles, range of motion, and phase-aware reasoning (see \emph{\textbf{Supp}} for more details).

\subsection{Results on QEVD-bio-fit-coach}

To assess biomechanics-grounded feedback on streaming video, we evaluate BioCoach on the new QEVD-bio-fit-coach benchmark and compare it to the strongest streaming baseline, Stream-VLM~\cite{panchal2024say}, which has been fine-tuned on the same annotations for fairness (Tab.~\ref{tab:results_enhanced}). This benchmark stresses anatomy/kinematics-centric language, quantitative ranges of motion, and phase-aware cues—precisely what safe, actionable coaching requires.

BioCoach markedly outperforms Stream-VLM across all metrics, with the largest relative gains in lexical measures (METEOR $+262.8\%$ and ROUGE-L $+179.6\%$), reflecting improved alignment with expert references. Judge-based evaluations also rise sharply: \emph{LLM-Bio-Acc.} shows the strongest improvement ($+89.5\%$) among judge metrics, indicating biomechanics-grounded, anatomy- and phase-accurate feedback. Temporal behavior remains strong (T-F-Score), showing that richer biomechanical grounding does not come at the expense of timely triggering. These gains stem from exercise-focused joint selection, personalized morphometrics, and cycle-aligned analysis that turn raw kinematics into verifiable cues, shifting the system from pattern matching to principled, auditable reasoning.


For a fair comparison of Stream-VLM~\cite{panchal2024say}, we re-train BioCoach on QEVD-fit-coach (with the original annotations), \emph{i.e.}, without any of our biomechanical labels, and evaluate it under the official protocol. This experiment serves two purposes: (i) it isolates architectural gains from annotation gains, showing that our biomechanics-grounded design helps even when trained with non-biomechanical labels; and (ii) it demonstrates \emph{backward compatibility}, indicating BioCoach can be deployed on datasets that lack fine-grained biomechanical supervision.
%
Under this setting (Tab.~\ref{tab:results_original}), BioCoach still surpasses the strongest fine-tuned baseline in text/semantic quality and judged correctness (METEOR $0.129$, ROUGE-L $0.122$, BERTScore $0.864$, LLM-Acc.\ $2.56$), while remaining near-parity in timing (T-F-Score $0.544$ vs.\ $0.56$). Together with the large improvements in QEVD-bio-fit-coach, these results show that (a) BioCoach improves quality under legacy annotations and (b) adding biomechanical supervision further unlocks its full potential for precise, anatomy-specific feedback.

\subsection{Ablation Study}
Ablations on QEVD-bio-fit-coach (Tab.~\ref{tab:ablation}) reveal a clear component hierarchy. Removing Motion Quality Context causes $\sim$57\% METEOR drop and LLM-Bio-Acc. decline (3.26 $\to$ 2.04), showing it is the core biomechanical driver. Removing Morphometric Context causes moderate degradation ($-9\%$ METEOR, $-6\%$ LLM-Bio-Acc), indicating that personalization improves specificity. DoF Selection provides modest gains ($-3.7\%$ LLM-Bio-Acc without it), filtering irrelevant kinematics. Reducing the temporal window to $2\,\mathrm{s}$ maintains text quality but drops the T-F-Score $\sim$24\%, confirming that $3\,\mathrm{s}$ offers better stability for cycle detection. The three modules divide labor: motion quality supplies \emph{what to say}, morphometrics supply \emph{for whom}, and DoF selection supplies \emph{where to look}—together enabling interpretable reasoning without sacrificing responsiveness.

\subsection{Limitations and Future Work}
Our approach depends on the quality of 3D skeletons and shapes, as occlusions, loose clothing, and extreme viewpoints can distort kinematics. Additionally, it relies on curated reference trajectories that may miss exercise variants and adaptive forms. Future work will explore multi-sensor fusion and broader reference coverage to mitigate these issues.
Beyond kinematics, we plan to extend BioCoach toward kinetic reasoning, estimating joint reaction forces and muscle activation patterns from video to detect compensatory movement strategies that are invisible to angle-only analysis. Integrating musculoskeletal simulation could further enable load-aware coaching that flags injury-risk forces, not just postural deviations
\section{Conclusion}

We present \textbf{BioCoach}, a biomechanics-grounded vision--language framework for streaming fitness coaching that exposes 3D kinematics, morphometrics, and constraint analyses to the language model. With an exercise-specific DoF selector, a morphometric and motion-quality context, and a vision--biomechanics conditioned feedback module, BioCoach delivers interpretable coaching rather than appearance-level heuristics. BioCoach improves lexical, semantic, and judge metrics while maintaining competitive timing, demonstrating that explicit kinematics and structured biomechanical context are key to accurate, phase-aware, personalized coaching. This formulation opens a path to coaching systems whose behavior can be inspected, trusted, and aligned with expert musculoskeletal reasoning.

\Paragraph{Acknowledgments} This work was supported in part by the National Science Foundation under Grant FMitF-2319242.



{
    \small
    \bibliographystyle{ieeenat_fullname}
    \bibliography{main}
}

\clearpage
\setcounter{page}{1}
\maketitlesupplementary

\begin{figure*}[t]
    \centering
    \includegraphics[width=0.95\linewidth]{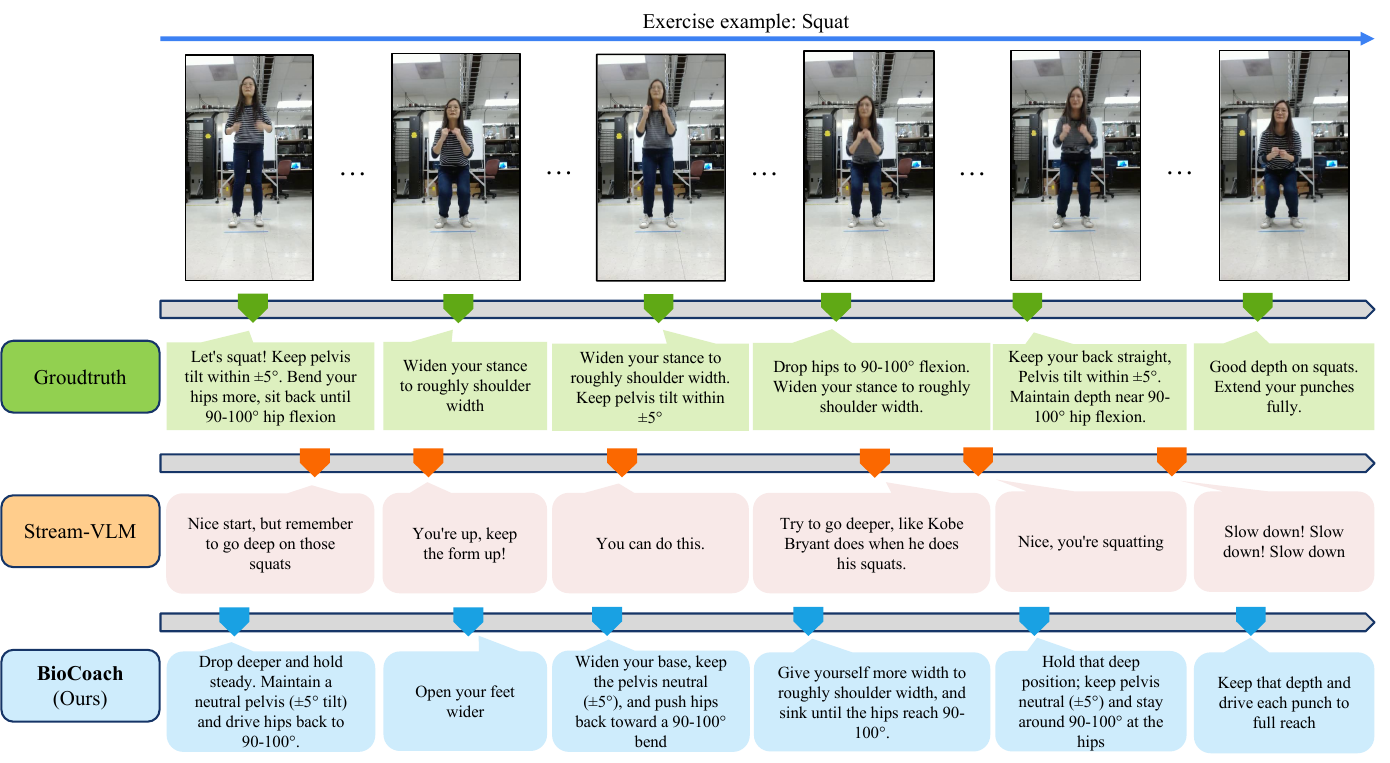}
    \caption{Qualitative timeline for a squat exercise. \textbf{BioCoach} produces temporally aligned, biomechanics-grounded cues with consistent phase tracking, while Stream-VLM outputs generic or mistimed feedback inconsistent with the ground-truth annotations.}
    \label{fig:supp_timeline}
\end{figure*}

This supplementary material provides the technical details referenced in the main paper. QEVD-bio-fit-coach dataset construction (Sec.~\ref{sec:dataset_construction}), evaluation metrics including the biomechanics-aware LLM judge (Sec.~\ref{sec:eval_metrics}), DoF annotation (Sec.~\ref{sec:dof_annotation_supp}), cycle detection (Sec.~\ref{sec:cycle_detection_details}), implementation specifics (Sec.~\ref{sec:impl_supp}) and mathematical derivations (Sec.~\ref{sec:math_derivations}).

\section{QEVD-bio-fit-coach Dataset Construction}
\label{sec:dataset_construction}
This section details the construction pipeline for QEVD-bio-fit-coach, which augments the original QEVD-fit-coach~\cite{panchal2024say} with biomechanically-grounded feedback annotations (referenced in the main paper Sec.~3.7).

\subsection{Automated Generation Pipeline}
\label{sec:dataset_pipeline}
The QEVD-bio-fit-coach dataset is created through an automated pipeline that applies the Motion Quality Context Module (Sec.~3.4.2 in the main paper) to transform colloquial coaching feedback into anatomically precise, quantified guidance.

\Paragraph{Selective Feedback Replacement.}
The original QEVD-fit-coach dataset contains diverse feedback types: corrective instructions addressing form deviations (\emph{e.g.}, ``lower your body more'', ``keep your back straight''), motivational encouragement (\emph{e.g.}, ``good job!'', ``keep it up!''), and repetition counting (\emph{e.g.}, ``5 more reps'').

We systematically rewrite \emph{corrective and instructional feedback}, as these inherently describe biomechanical violations that can be quantified. At each corrective feedback timestamp, the trainer observed a form deviation; our pipeline detects and quantifies the actual biomechanical violation, then replaces the colloquial description with precise measurements. Motivational and counting feedback are retained unchanged, as they do not describe specific form issues.

\Paragraph{Generation Procedure.}
For each video in the original QEVD-fit-coach dataset, we apply the complete BioCoach pipeline to generate biomechanical context:

\begin{enumerate}[leftmargin=*,itemsep=1pt]
\item Extract 3D skeletal kinematics using HSMR + SKEL (Sec.~\ref{sec:skeletal_supp}) and morphometric measurements (main paper Sec.~3.4.1)

\item Apply the trained DoF selection network $\mathcal{A}_\theta$ (Sec.~\ref{sec:dof_annotation_supp}) to identify exercise-specific salient joints $\mathcal{J}^*$

\item At each corrective/instructional feedback timestamp, run the Motion Quality Context Module pipeline (main paper Sec.~3.4.2): cycle detection, reference alignment, and biomechanical constraint evaluation to generate structured motion context $\mathcal{C}_{\text{motion}}$

\item Replace the corrective/instructional feedback with the automatically generated motion context, which provides:
\begin{itemize}[leftmargin=*,itemsep=1pt]
    \item Quantified form violations (\emph{e.g.}, ``Right knee flexion 85$^{\circ}$, target 90$^{\circ}$'')
    \item Anatomically precise corrective instructions (\emph{e.g.}, ``Increase knee flexion by 5$^{\circ}$ at the bottom phase'')
\end{itemize}
\end{enumerate}

The timestamp of each feedback instance is preserved exactly from the original QEVD-fit-coach; only the feedback content is replaced with biomechanically-grounded descriptions.

\subsection{Dataset Statistics}
\label{sec:dataset_stats}

The resulting QEVD-bio-fit-coach dataset maintains an identical structure to the original QEVD-fit-coach:
\begin{itemize}[leftmargin=*,itemsep=1pt]
\item \textbf{Training set}: $149$ videos across $23$ exercise types
\item \textbf{Test set}: 74 videos across the same 23 exercise types
\end{itemize}

\Paragraph{Annotation Examples.}
Representative transformations from the original to biomechanically-grounded feedback:
\begin{itemize}[leftmargin=*,itemsep=1pt]
\item \textbf{Squat}: ``Lower your body more'' $\to$ ``Knee flexion 85$^{\circ}$ (target: 90$^{\circ}$). Increase knee flexion by 5$^{\circ}$ at the bottom phase.''
\item \textbf{Push-up}: ``Keep your back straight'' $\to$ ``Lumbar spine variance 12$^{\circ}$ (target: $<5^{\circ}$). Maintain neutral spine alignment; engage the core to stabilize the lumbar region.''
\item \textbf{Lunge}: ``Don't lean forward'' $\to$ ``Thorax forward lean 23$^{\circ}$ from vertical. Reduce thorax flexion by 8$^{\circ}$; keep the torso upright.''
\item \textbf{Plank}: ``Hold steady'' $\to$ ``Hip angle variance 8$^{\circ}$ (target: $<5^{\circ}$). Stabilize the hip position and maintain static alignment from the shoulders to the ankles.''
\end{itemize}

This automated pipeline ensures consistent, anatomically precise feedback grounded in explicit biomechanical analysis, enabling systematic evaluation of biomechanics-aware coaching systems.

\section{Evaluation Metrics}
\label{sec:eval_metrics}
We summarize the LLM-based automatic scoring pipelines; each metric uses a distinct prompt and decoding setup.


\Paragraph{LLM-Bio-Accuracy}
\begin{itemize}[leftmargin=*,itemsep=1pt]
\item \textbf{Biomech system prompt}: ``You are an expert biomechanics analyst. Your role is to evaluate the factual accuracy and relevance of the biomechanical feedback generated for a user's exercise performance. Always respond as a python dictionary string.''
\item \textbf{Biomech user template}:
\texttt{Please evaluate the following predicted biomechanical feedback:}\\
\texttt{-Ground truth biomechanical context: <1>}\\
\texttt{-Predicted biomechanical feedback: <2>}\\
\texttt{Compare the predicted biomechanical feedback against the ground truth context. Focus}\\
\texttt{on maintaining factual accuracy regarding joint angles, posture, balance, and other factors.}\\
\texttt{biomechanical observations. Provide an integer accuracy score between 1 and 5}\\
\texttt{(5 = perfect alignment). Respond strictly as a Python dictionary string:}\\
\texttt{\{'score': int(score)\} with no extra text.}
\end{itemize}


\section{Exercise-Specific DoF Annotation Protocol}
\label{sec:dof_annotation_supp}

Ground-truth salient joint sets $\mathcal{J}^{\text{gt}}$ were obtained using Qwen3-VL-30B-A3B-Instruct with biomechanics domain knowledge. Crucially, annotations are conditioned on both \emph{exercise type} and \emph{execution quality}. For each video segment, Qwen3-VL-30B-A3B-Instruct was provided with the exercise type (\emph{e.g.}, squat) and the video frames. The vision-language model then identified which joints require attention by analyzing visual cues of execution quality (\emph{e.g.}, detecting forward lean). Domain experts reviewed and validated these LLM-generated annotations. For example:
\begin{itemize}[leftmargin=*,itemsep=1pt]
\item Standard salient joints for squat = \{hips, knees, ankles, lumbar\}
\item Expanded set for squat with forward lean = \{hips, knees, ankles, lumbar, thorax\}
\end{itemize}

This quality-aware annotation strategy enables the DoF selection network $\mathcal{A}_\theta$ to learn \emph{execution-dependent} attention patterns. Unlike fixed per-exercise lookup tables, the learned selector can identify form deviations from visual cues and dynamically prioritize relevant joints, even when exercise variants or error patterns are not explicitly labeled. Similar quality-dependent annotations were obtained for all 23 exercises, with an average of $K=10$--15 joints per video segment depending on detected form deviations.

\section{Static vs. Dynamic Joint Classification}
\label{sec:static_dynamic_supp}
As referenced in Sec.~3.4.2 (``Biomechanical Constraint Evaluation''), each joint is classified as \emph{static} (should remain stable) or \emph{dynamic} (should exhibit specific motion patterns).

\Paragraph{Classification Procedure.}
For each of the 23 exercise types, GPT-4 was used to classify joints as static or dynamic. The model was prompted with exercise descriptions and biomechanical principles to determine, for each joint, whether it should maintain stability or exhibit controlled motion during that exercise. This produces a per-exercise lookup table of joint classifications. Domain experts reviewed and validated these LLM-generated annotations. At inference time, biomechanical constraint evaluation (Sec.~3.4.2) is applied only to the intersection of the DoF selector's predicted salient joints $\mathcal{J}^*$.
\begin{itemize}[leftmargin=*,itemsep=1pt]
\item \textbf{Static joints}: Should maintain stable angles throughout the cycle. Evaluated via variance: $\delta_j^{\text{static}} = \text{Var}(\{q_{j,i}\}_{i \in [i_s, i_e]})$. Example: spine alignment during squats should have $\delta_j^{\text{static}} < 5^{\circ}$
\item \textbf{Dynamic joints}: Should follow specific motion patterns and reach target ranges at critical frames. Evaluated via deviation from reference at key frames: $\delta_j^{\text{dynamic}} = |q_j^{\text{user}}(i_{\text{key}}) - q_j^{\text{ref}}(i_{\text{key}})|$. Example: hip flexion during squats should reach $90^{\circ}$--$100^{\circ}$ at the bottom position
\end{itemize}

\Paragraph{Exercise-Specific Examples.}
Below, we provide complete joint classifications for all 24 anatomical joints across representative exercises:
\begin{itemize}[leftmargin=*,itemsep=1pt]
\item \textbf{Squat}:
\begin{itemize}[leftmargin=*,itemsep=0pt]
\item Dynamic: R/L hip, R/L knee, R/L ankle, R/L subtalar
\item Static: pelvis, lumbar, thorax, head, R/L scapula, R/L shoulder, \ldots (remaining upper-body joints)
\end{itemize}

\item \textbf{Push-up}:
\begin{itemize}[leftmargin=*,itemsep=0pt]
\item Dynamic: R/L scapula, R/L shoulder, R/L elbow
\item Static: pelvis, lumbar, thorax, head, R/L hip, R/L knee, \ldots (remaining lower-body joints)
\end{itemize}

\item \textbf{Plank}:
\begin{itemize}[leftmargin=*,itemsep=0pt]
\item Dynamic: none (isometric hold)
\item Static: all 24 joints (pelvis, lumbar, thorax, head, R/L hip, knee, ankle, \ldots)
\end{itemize}

\item \textbf{Lunge}:
\begin{itemize}[leftmargin=*,itemsep=0pt]
\item Dynamic: R/L hip, R/L knee, R/L ankle
\item Static: pelvis, lumbar, thorax, head, R/L scapula, R/L shoulder, \ldots (remaining joints)
\end{itemize}
\end{itemize}

\section{Cycle Detection Algorithm Details}
\label{sec:cycle_detection_details}
This section provides implementation parameters for the cycle detection algorithm described in the main paper, Sec.~3.4.2.

\Paragraph{Video Sampling and Preprocessing.}
Original videos are captured at 30\,fps. Gaussian smoothing is applied using \texttt{scipy.ndimage.gaussian\_filter1d} with $\sigma=2$.

\Paragraph{Representative Joint Selection.}
For each exercise type, we select a representative joint whose angle trajectory exhibits the clearest periodic pattern for cycle detection. The joint is chosen empirically: squats utilize knee flexion, and push-ups involve elbow flexion. For alternating movements, we select central joints that reflect bilateral patterns (high knees use pelvis tilt), enabling cycle detection and subsequent left-right phase segmentation.

\Paragraph{Prominence-Based Peak Detection Parameters.}
For repetitive exercises, we use \texttt{scipy.signal.find\_peaks} with:
\begin{itemize}[leftmargin=*,itemsep=1pt]
\item Prominence threshold: $p_{\text{min}} = 0.1$ radians ($\approx 5.7^{\circ}$)
\item Distance constraint: $d_{\text{min}} = 5$ frames (half of minimum cycle length)
\item Cycle length bounds: $[24, 150]$ frames (0.8--5.0\,s at 30\,fps)
\end{itemize}

\Paragraph{Alternating Movement Detection Parameters.}
For alternating movements (\emph{e.g.}, high knees), after detecting cycle boundaries $(i_s, i_e)$ via prominence-based peak detection on the central joint trajectory, zero-crossing detection is applied within the cycle to identify phase transitions. We compute the cycle-specific mean:
\begin{equation}
\bar{q}_{j,\text{cycle}} = \frac{1}{i_e - i_s + 1}\sum_{i=i_s}^{i_e} q_{j,i},
\end{equation}
and detect zero-crossings of the normalized signal $q_{j,i} - \bar{q}_{j,\text{cycle}}$ to segment left-right alternation phases. These phase boundaries enable the evaluation of bilateral limb patterns: left and right limb angles can be compared within their respective phases to assess symmetry.

\Paragraph{Static Hold Detection Parameters.}
For isometric holds, rolling variance is computed with a window of $w = 10$ frames. Stability threshold $\epsilon = P_{30}(\{\sigma_t\})$ (30th percentile of rolling standard deviations). Stable segments $\geq 10$ frames are extracted as hold cycles.

\Paragraph{Reference Trajectory Source.}
Reference trajectories are curated using exercise-specific biomechanical rules: for each of the 23 exercises, we define canonical joint angle ranges and motion patterns based on sports science literature and domain expertise. These rules generate complete 46-dimensional angle representations with natural within-cycle variations, saved as \texttt{.npy} files. At inference, these pre-computed references serve as baselines for reference alignment and constraint evaluation. 

\Paragraph{Cycle Quality Score Computation.}
The four similarity metrics mentioned in main paper Sec.~3.4.2 are computed as follows: (1) all salient joint angles are concatenated and z-score normalized before computing cosine similarity; (2) Pearson correlations are averaged across salient joints $\mathcal{J}^*$; (3) frame-to-frame differences $\Delta q_{j,k} = q_{j,k+1} - q_{j,k}$ are computed for velocity comparison; (4) range of motion $\text{ROM}_j = \max(q_j) - \min(q_j)$ ratios are converted to $[0,1]$ and averaged. The final score combines these via weighted sum: $s_{\text{cycle}} = 0.4 \cdot \text{sim}_{\text{cos}} + 0.3 \cdot \text{sim}_{\text{pearson}} + 0.2 \cdot \text{sim}_{\text{vel}} + 0.1 \cdot \text{sim}_{\text{amp}}$, clipped to $[0,1]$. This score provides cycle quality confidence and enables best-cycle selection when multiple candidates exist.

\section{Implementation Details}
\label{sec:impl_supp}
This section expands on the implementation overview in the main paper (Sec.~3.8), providing architecture specifications and training configurations.

\subsection{DoF Selection Network Architecture}
\label{sec:dof_network_arch}
The attention network $\mathcal{A}_\theta$ (referenced in Sec.~3.3 and trained as described in Sec.~3.6) is a 3-layer MLP with learnable parameters $\theta$ that takes visual features $\mathbf{F}_t^{\text{vis}}$ as input and outputs importance scores for each joint. The network consists of:
\begin{itemize}[leftmargin=*,itemsep=1pt]
\item \textbf{Layer 1}: Projects pooled visual tokens to an intermediate 512-dimensional space with ReLU activation and a dropout rate of 0.1
\item \textbf{Layer 2}: 256-dimensional hidden layer with ReLU activation and dropout rate of 0.1
\item \textbf{Layer 3}: Output layer producing $J=24$ importance scores $\mathbf{s}^t \in [0,1]^{24}$ via sigmoid activation
\item \textbf{Top-K selection}: Selects top $K=12$ joints by importance for downstream biomechanical analysis
\end{itemize}
\subsection{Visual Backbone Configuration}
\label{sec:visual_backbone_supp}
As referenced in the main paper (Sec.~3.2, paragraph ``Visual Appearance Backbone''), we employ a 3D CNN following~\cite{panchal2024say}.

\Paragraph{Visual Appearance Backbone Details.}
Temporal window: $\tau=12$ frames sampled at 4 Hz from the $30$\,fps video ($3$\,s motion history), where each frame represents a temporal sampling point processed by the 3D CNN

Output tokens: $N_v=35$ visual tokens with embedding dimension $d=1280$

Architecture: 2D and 3D convolutional layers with causal masking (detailed in the main paper, Sec.~3.2). All 3D CNN weights are frozen during training; only the cross-attention layers (Sec.~\ref{sec:crossattn_derivation}) are trainable

\subsection{3D Skeletal Extraction and Morphometric Processing}
\label{sec:skeletal_supp}
We extract biomechanically-grounded skeletal kinematics using HSMR~\cite{xia2025reconstructing} + SKEL~\cite{keller2023skin} (referenced in Sec.~3.2, ``3D Skeletal Kinematic Backbone'').

\Paragraph{46-Dimensional Skeletal Representation.}
Following SKEL~\cite{keller2023skin}, we use a 46-dimensional Euler-angle representation that defines 24 anatomical joints. The degrees of freedom (DoFs) are organized as:
\begin{itemize}[leftmargin=*,itemsep=1pt]
\item \textbf{Pelvis} (0--2): tilt, list, rotation
\item \textbf{Right leg} (3--9): hip (flexion, adduction, rotation), knee angle, ankle angle, subtalar angle, mtp angle
\item \textbf{Left leg} (10--16): hip (flexion, adduction, rotation), knee angle, ankle angle, subtalar angle, mtp angle
\item \textbf{Spine} (17--25): lumbar (bending, extension, twist), thorax (bending, extension, twist), head (bending, extension, twist)
\item \textbf{Right arm} (26--35): scapula (abduction, elevation, upward rotation), shoulder (x, y, z), elbow flexion, pro/sup, wrist (flexion, deviation)
\item \textbf{Left arm} (36--45): scapula (abduction, elevation, upward rotation), shoulder (x, y, z), elbow flexion, pro/sup, wrist (flexion, deviation)
\end{itemize}

\Paragraph{3D Skeletal Kinematic Backbone Details.}
Skeletal representation: 46-dimensional Euler-angle representations with joint-specific biomechanical constraints following SKEL~\cite{keller2023skin}

Temporal aggregation: Shape parameters $\boldsymbol{\beta}$ averaged over the $\tau=12$ frame window to yield stable $\bar{\boldsymbol{\beta}}$, reducing per-frame shape estimation noise

Joint angle smoothing: Gaussian smoothing is applied to joint angle trajectories to reduce jitter while preserving cycle peaks for detection

Morphometric conversion: Virtual Measurements~\cite{choutas2022accurate} extracts \texttt{mass}, \texttt{height}, \texttt{chest}, \texttt{waist}, and \texttt{hip} circumference from the fitted SMPL mesh (procedure detailed in the main paper Sec.~3.4.1)

Confidence handling: When HSMR confidence is low (severe occlusion, extreme angles), the system retains the last valid pose and prepends warnings to $\mathcal{C}_{\text{morph}}$

\subsection{Training Configuration}
\label{sec:training_supp}
As noted in Sec.~3.6, we employ a fine-tuning strategy that freezes both the 3D CNN visual backbone and the LLaMA-2-7B language model. Training proceeds in two stages: first, training the DoF selection network $\mathcal{A}_\theta$, then fixing it and training the cross-attention fusion layers.

\Paragraph{Stage 1: DoF Selection Network Pre-training.}
The DoF selection network $\mathcal{A}_\theta$ is first trained independently using binary cross-entropy loss $\mathcal{L}_{\text{DoF}}$ (Eq. in main paper Sec.~3.6) with ground-truth salient joint annotations $\mathcal{J}^{\text{gt}}$ obtained via Qwen3-VL-30B (detailed in Sec.~\ref{sec:dof_annotation_supp}). The network is initialized randomly and trained until the joint selection accuracy converges. Once trained, $\mathcal{A}_\theta$ is frozen to provide consistent, salient joint predictions during subsequent cross-attention training.

\Paragraph{Stage 2: Cross-Attention Fusion Training.}
With $\mathcal{A}_\theta$ frozen, cross-attention projection matrices $W_Q, W_K, W_V, W^O$ are trained via autoregressive cross-entropy loss $\mathcal{L}_{\text{CE}}$ with action-token down-weighting (Eq. in main paper Sec.~3.6). Initialized with Xavier uniform (gain${}\!=\!0.1$), the layers learn to fuse visual and morphometric features while the visual and LLM backbones remain frozen.


\section{Mathematical Derivations}
\label{sec:math_derivations}
This section expands the mathematical formulations for vision-biomechanics conditioning (referenced in Sec.~3.5).

\subsection{Vision-Morphometric Cross-Attention Details}
\label{sec:crossattn_derivation}
The cross-attention mechanism (Eq.~\eqref{eq:crossattn} in the main paper) fuses visual features $\mathbf{F}_t^{\text{vis}} \in \mathbb{R}^{N_v \times d}$ with morphometric context $\mathbf{m}_t \in \mathbb{R}^{N_m \times d}$.

\Paragraph{Architecture Details.}
\begin{itemize}[leftmargin=*,itemsep=1pt]
\item \textbf{Multi-head configuration}: $h=8$ attention heads with per-head dimension $d_k=160$ 
\item \textbf{Initialization}: Cross-attention projection matrices $W_Q, W_K, W_V, W^O$ initialized with Xavier uniform (gain${}\!=\!0.1$)
\item \textbf{Residual connection}: $\mathbf{z}_t = \mathbf{F}_t^{\text{vis}} + \text{CrossAttn}(\cdot)$ prevents morphometric context from overriding visual evidence
\end{itemize}

\subsection{Motion Context Serialization and Prompting}
\label{sec:context_encoding_supp}
Motion quality context $\mathcal{C}_{\text{motion}}$ (Sec.~3.4.2 in the main paper) is serialized as structured natural language, combining pose state and detected violations.

\Paragraph{Continuous Cycle Detection and Motion Context Updates.}
During streaming inference, cycle detection runs continuously: as frames arrive, the angle trajectory is updated, and the peak detection algorithm incrementally identifies completed cycles. At each timestep, the motion quality context $\mathcal{C}_{\text{motion}}$ is generated by combining (1) the current pose state $\mathbf{p}_{\text{state}}^{(t)}$, which captures instantaneous joint angles, and (2) violations $\mathbf{v}_{\text{violations}}$ from the most recently completed cycle's constraint evaluation (Sec.~3.4.2). The pose state updates every frame, while violations are updated whenever a new cycle completes. This $\mathcal{C}_{\text{motion}}$ is prepended to the prompt at each token generation step (main paper Eq. in Sec.~3.5), ensuring the LLM always conditions on up-to-date biomechanical analysis while maintaining streaming operation.






\end{document}